\documentclass[10pt,journal,compsoc]{IEEEtran}

\usepackage{amsmath}
\usepackage{amssymb,subfigure,cases}
\usepackage{color,hyperref}
\usepackage{algpseudocode}
\usepackage{booktabs}

\usepackage{amsmath}
\usepackage{subfigure}
\usepackage{caption}
\usepackage{algorithm}
\usepackage{algpseudocode}
\usepackage{setspace}

\usepackage{soul}
\usepackage{pifont}

\usepackage{times}
\usepackage{epsfig}
\usepackage{graphicx}
\usepackage{amssymb}
\usepackage{multirow}
\usepackage{subfigure}
\usepackage{pifont}
\usepackage{color,xcolor,colortbl}
\usepackage{array}
\usepackage{url}
\usepackage{CJK}
\usepackage{makecell}

\definecolor{mygray}{gray}{.9}

\newcommand{\xmark}{\ding{55}}
\newcommand{\rmark}{\ding{52}}

\newcolumntype{I}{!{\vrule width 1.2pt}}
\newlength\savedwidth
\newcommand\whline{\noalign{\global\savedwidth\arrayrulewidth
		\global\arrayrulewidth 1.25pt}%
	\hline
	\noalign{\global\arrayrulewidth\savedwidth}}

\ifCLASSOPTIONcompsoc
  \usepackage[nocompress]{cite}
\else
  \usepackage{cite}
\fi

\ifCLASSINFOpdf
\else
 \fi

\begin{document}
	
\setul{}{1.0pt}
%
\title{NWPU-Crowd: A Large-Scale Benchmark for Crowd Counting and Localization}

\author{
     Qi~Wang, \IEEEmembership{Senior~Member,~IEEE,} Junyu Gao, \IEEEmembership{Student~Member,~IEEE,} Wei Lin, \\Xuelong Li*, \IEEEmembership{Fellow,~IEEE}
\IEEEcompsocitemizethanks{
\IEEEcompsocthanksitem
\textit{Xuelong Li is the corresponding author.}
\IEEEcompsocthanksitem
Qi Wang, Junyu Gao, Wei Lin, and Xuelong Li are with the School of Computer Science and with the Center for Optical Imagery Analysis and Learning (OPTIMAL), Northwestern Polytechnical University, Xi'an 710072, Shaanxi, China. E-mails: crabwq@gmail.com, gjy3035@gmail.com, elonlin24@gmail.com, li@nwpu.edu.cn. 
\IEEEcompsocthanksitem
\copyright 20XX IEEE. Personal use of this material is permitted. Permission from IEEE must be obtained for all other uses, in any current or future media, including reprinting/republishing this material for advertising or promotional	purposes, creating new collective works, for resale or redistribution to servers or lists, or reuse of any copyrighted component of this work in other works.
}}

\markboth{IEEE TRANSACTIONS ON PATTERN ANALYSIS AND MACHINE INTELLIGENCE,~Vol.~XXX, No.~XXX, XXX~XXX}%
{Shell \MakeLowercase{\textit{et al.}}: Bare Advanced Demo of IEEEtran.cls for IEEE Computer Society Journals}

\IEEEtitleabstractindextext{%
\begin{abstract}

In the last decade, crowd counting and localization attract much attention of researchers due to its wide-spread applications, including crowd monitoring, public safety, space design, etc. Many Convolutional Neural Networks (CNN) are designed for tackling this task. However, currently released datasets are so small-scale that they can not meet the needs of the supervised CNN-based algorithms. To remedy this problem, we construct a large-scale congested crowd counting and localization dataset, NWPU-Crowd, consisting of $5,109$ images, in a total of $2,133,375$ annotated heads with points and boxes. Compared with other real-world datasets, it contains various illumination scenes and has the largest density range ($0\!\sim\!20,033$). Besides, a benchmark website is developed for impartially evaluating the different methods, which allows researchers to submit the results of the test set. Based on the proposed dataset, we further describe the data characteristics, evaluate the performance of some mainstream state-of-the-art (SOTA) methods, and analyze the new problems that arise on the new data. What's more, the benchmark is deployed at \url{https://www.crowdbenchmark.com/}, and the dataset/code/models/results are available at \url{https://gjy3035.github.io/NWPU-Crowd-Sample-Code/}. 
\end{abstract}

}

\maketitle

\IEEEpeerreviewmaketitle

\section{Introduction}
\label{intro}

\IEEEPARstart{C}{rowd} analysis is an essential task in the field of video surveillance. Accurate analysis for crowd motion, human behavior, population density is crucial to public safety, urban space design, \emph{etc.} Crowd counting and localization are fundamental tasks in the field of crowd analysis, which serve high-level tasks, such as crowd flow estimation \cite{ali2007lagrangian} and pedestrian tracking \cite{yu2016solution}. Due to the importance of crowd counting, many researchers \cite{sindagi2017generating,liu2019exploiting,wan2019residual} pay attention to it and achieve quite a few significant improvements in this field. Especially, benefiting from the development of deep learning in computer vision, the counting performance on the datasets \cite{chan2008privacy,zhang2016single,zhang2016data,idrees2018composition} is continuously refreshed by Convolutional Neural Networks (CNN)-based methods \cite{li2018csrnet,ranjan2018iterative,liu2019crowd}. 

The CNN-based methods need to learn discriminate features from a multitude of labeled data, so a large-scale dataset can effectively promote the development of visual technologies. It is verified in many existing tasks, such as object detection \cite{lin2014microsoft} and semantic segmentation \cite{neuhold2017mapillary}. However, the currently released crowd counting datasets are so small-scale that most deep-learning-based methods are prone to overfit the data. According to the statistics, UCF-QNRF \cite{idrees2018composition} is the largest released congested crowd counting dataset. Still, it contains only $1,535$ samples, in a total of $1.25$ million annotated instances, which is still unable to meet the needs of current deep learning methods. Moreover, some works \mbox{\cite{idrees2018composition,liu2019recurrent}} focus on the crowd localization task that produces point-wise predictions for each instance. However, the traditional datasets do not contain box-level labels, which makes it hard to evaluate the localization performance using a uniform metric. Furthermore, there is not an impartial evaluation benchmark, which potentially restricts further development of crowd counting. By the way, some methods\footnote{https://github.com/gjy3035/Awesome-Crowd-Counting/issues/78} may use mistaken labels to evaluate models, which is also not accurate. Reviewing some benchmarks in other fields, CityScapes \cite{cordts2016cityscapes} and Microsoft COCO \cite{lin2014microsoft}, they allow the researchers to submit their results of the test set and impartially evaluate them, which facilitates the study of methodology. Thus, an equitable evaluation platform is important for the community.

Considering the problems mentioned above, in this paper, we construct a large-scale crowd counting and localization dataset, named as \textbf{NWPU-Crowd}, and develop a benchmark website to boost the community of crowd analysis. Compared with the existing congested datasets, the proposed NWPU-Crowd has the following main advantages: 1) This is the largest crowd counting and localization dataset, consisting of $5,109$ images and containing $2,133,375$ annotated instances; 2) It introduces some negative samples like high-density crowd images to assess the robustness of models; 3) In NWPU-Crowd, the number of annotated objects range, $0\sim20,033$. More concrete features are described in Section~\ref{characteristic}. Table~\mbox{\ref{data-com}} illustrates the detailed statistics of ten mainstream real-world datasets and the proposed NWPU-Crowd. 

\begin{table*}[htbp]	
	\scriptsize
	\centering
	\caption{Statistics of the ten mainstream crowd counting datasets and NWPU-Crowd.}
	\setlength{\tabcolsep}{0.9mm}{
	\begin{tabular}{c|c|c|c|c|c|c|c|c|c|c}
		\whline
		\multirow{2}{*}{Dataset}	&Number &Avg. Resolution &\multicolumn{4}{c|}{Count Statistics} &Extreme &Unseen & Category-wise &Box-level \\
		\cline{4-7} 
		& of Images & ($H \times W$) & Total &Min & Ave & Max &Congestion &Test Labels & Evaluation	&Label\\
		\whline
		UCSD \cite{chan2008privacy}   &2,000 &$158 \times 238$  & 49,885 &11 & 25 & 46 &\xmark &\xmark &\xmark &\xmark	\\
		\hline
		Mall  \cite{chen2012feature}  &2,000 &$480 \times 640$  & 62,325 &13 & 31 & 53 &\xmark &\xmark &\xmark &\xmark	\\
		\hline
		WorldExpo'10 \cite{zhang2016data} &3,980 &$576 \times 720$  & 199,923 &1 & 50 & 253 &\xmark &\xmark &\rmark &\xmark \\
		\hline
		ShanghaiTech Part B \cite{zhang2016single}  &716 &$768 \times 1024$  & 88,488 &9 & 123 & 578 &\xmark &\xmark &\xmark &\xmark \\
		\hline
		Crowd\_Surv \cite{yan2019perspective} &13,945 &$840 \times  1342$  & 386,513 &2 & 35 & 1,420 &\xmark &\xmark&\xmark &\xmark \\
		\whline
		UCF\_CC\_50  \cite{idrees2013multi} &50 &$2101 \times 2888$  & 63,974 &94 & 1,279 & 4,543 &\rmark &\xmark &\xmark &\xmark	\\
		\hline
		ShanghaiTech Part A \cite{zhang2016single}  &482 &$589 \times 868 $  & 241,677 &33 & 501 & 3,139 &\rmark &\xmark &\xmark &\xmark \\
		\hline
		UCF-QNRF \cite{idrees2018composition} &1,535 &$2013 \times 2902$  & 1,251,642 & 49 & 815 & 12,865 &\rmark &\xmark &\xmark &\xmark\\
		\hline	
		GCC (synthetic) \mbox{\cite{wang2019learning}} &15,212 &$1080\times1920$  & 7,625,843 & 0 & 501 & 3,995 &\rmark &\xmark &\rmark &\xmark \\
		\hline		
		JHU-CROWD++ \cite{sindagi2019pushing,sindagi2020jhu}   &4,372 &$910 \times 1430$  & 1,515,005 & 0 & 346 & 25,791 &\rmark &\xmark  &\rmark &\rmark\\
		\whline
		\textbf{NWPU-Crowd}  &\textbf{5,109} &$\boldsymbol{2191 \times 3209}$  &\textbf{2,133,375}  &\textbf{0} &\textbf{418} &\textbf{20,033} &\rmark & \rmark &\rmark &\rmark \\ 
		\whline
		
	\end{tabular}}\label{data-com}
\end{table*}

Based on the proposed NWPU-Crowd, several experiments of some classical and state-of-the-art methods are conducted. After further analyzing their results, an interesting phenomenon on the proposed dataset is found: diverse data makes it difficult for counting networks to learn useful and distinguishable features, which does not appear or is ignored in the previous datasets. Specifically, 1) there are many error estimations on negative samples; 2) the data of different scene attributes (density level and luminance) have a significant influence on each other. Therefore, it is a research trend on how to alleviate the above two problems. What's more, for localization task, we design a reasonable metric and provide some simple baseline models.

In summary, we believe that the proposed large-scale dataset will promote the application of crowd counting and localization in practice and attract more attention to tackling the aforementioned problems.

\section{Related Works}
\label{related}

The existing crowd counting datasets mainly contain two types: surveillance-scene datasets and general-scene datasets. The former commonly records crowd in particular scenarios, of which the data consistency is obvious. For the latter, the crowd samples are collected from the Internet. Thus, there are more perspective variations, occlusions, and extreme congestion in these datasets. Tabel~\ref{data-com} demonstrates a summary of the basic information of the mainstream crowd counting datasets, and in the following parts, their unique characters are briefly introduced. 

\subsection{Surveillance-scene Dataset} 

\textbf{Surveillance view. } Surveillance-view datasets aim to collect the crowd images in specific indoor scenes or small-area outdoor locations, such as marketplace, walking street, and station. The number of people usually ranges from 0 to 600. UCSD is a typical dataset for crowd analysis. It contains $2,000$ image sequences, which records a pedestrian walk-way at the University of California at San Diego (UCSD). Mall~\cite{chen2012feature} is captured in a shopping mall with more perspective distortion. However, these two datasets contain only a single scene, lacking data diversity. Thus, Zhang \emph{et al.} \cite{zhang2016data} build a multi-scene crowd counting dataset, WorldExpo'10, consisting of 108 surveillance cameras with different locations in Shanghai 2010 WorldExpo, e.g., entrance, ticket office. Considering the poor resolution of traditional surveillance cameras, Zhang \emph{et al.} \cite{zhang2016single} construct a high-quality crowd dataset, ShanghaiTech Part B, containing $782$ images captured in some famous resorts of Shanghai, China. To remedy the occlusion problem in congested scenes, a multi-view dataset is designed by Zhang and Chan \cite{zhang2019wide}. By equipping 5 cameras at different positions for a specific view, the data can be recorded synchronously. For getting rid of the manually labeling process, Wang \emph{et al.} \cite{wang2019learning} construct a large-scale synthetic dataset (GCC). By simulating the perspective of a surveillance camera, they capture $400$ crowd scenes in a computer game (Grand Theft Auto V, GTA V), a total of $15,212$ images. The main advantage of GCC is that it can provide accurate labels (point and mask) and diverse environments. However, there are many domain shifts/gaps between synthetic and real data, limiting their practical values. Therefore, it is necessary to build a large-scale real-world dataset. Compared with GCC, the advantages of NWPU-Crowd are: more natural person models, crowd scenes and environment (weathers, light, etc.).

In addition to the aforementioned datasets, there are also other crowd counting datasets with their specific characteristics. 
SmartCity~\cite{zhang2018crowd} focuses on some typical scenes, such as sidewalk and subway. ShanghaiTechRGBD~\cite{Lian_2019_CVPR} records the RGBD crowd images with a stereo camera for concentrating on pedestrian counts and localization. Fudan-ShanghaiTech \cite{fang2019locality} and Venice \cite{Liu_2019_CVPR} capture the video sequences for temporal crowd counting.

\noindent\textbf{Drone view. } For some big scenes (such as stadium, plaza) or some large rally events (ceremony, hajj, \emph{etc.}), the above traditional fixed surveillance camera is not suitable due to its small field of view. To tackle this problem, some other datasets are collected through the Drone or Unmanned Aerial Vehicle (UAV). Benefiting from their higher altitudes, more flexible view and free flight, more large scenes can be recorded compared with the traditional surveillance camera. There are two crowd counting datasets with the drone view, DLR-ACD Dataset~\cite{bahmanyar2019mrcnet} and DroneCrowd Dataset~\cite{wen2019drone}. The former consists of 33 images with $226,291$ annotated persons, including some mass events: sports, concerts, trade fair, \emph{etc.} The latter consists of $70$ crowd scenes , with a total of $33,600$ drone-view image sequences. Due to the Bird's-Eye View (BEV), the whole body of pedestrians can not be seen except their heads, so the perspective change rarely appears in the above two datasets.

\begin{figure*}
	\centering
	\includegraphics[width=0.8\textwidth]{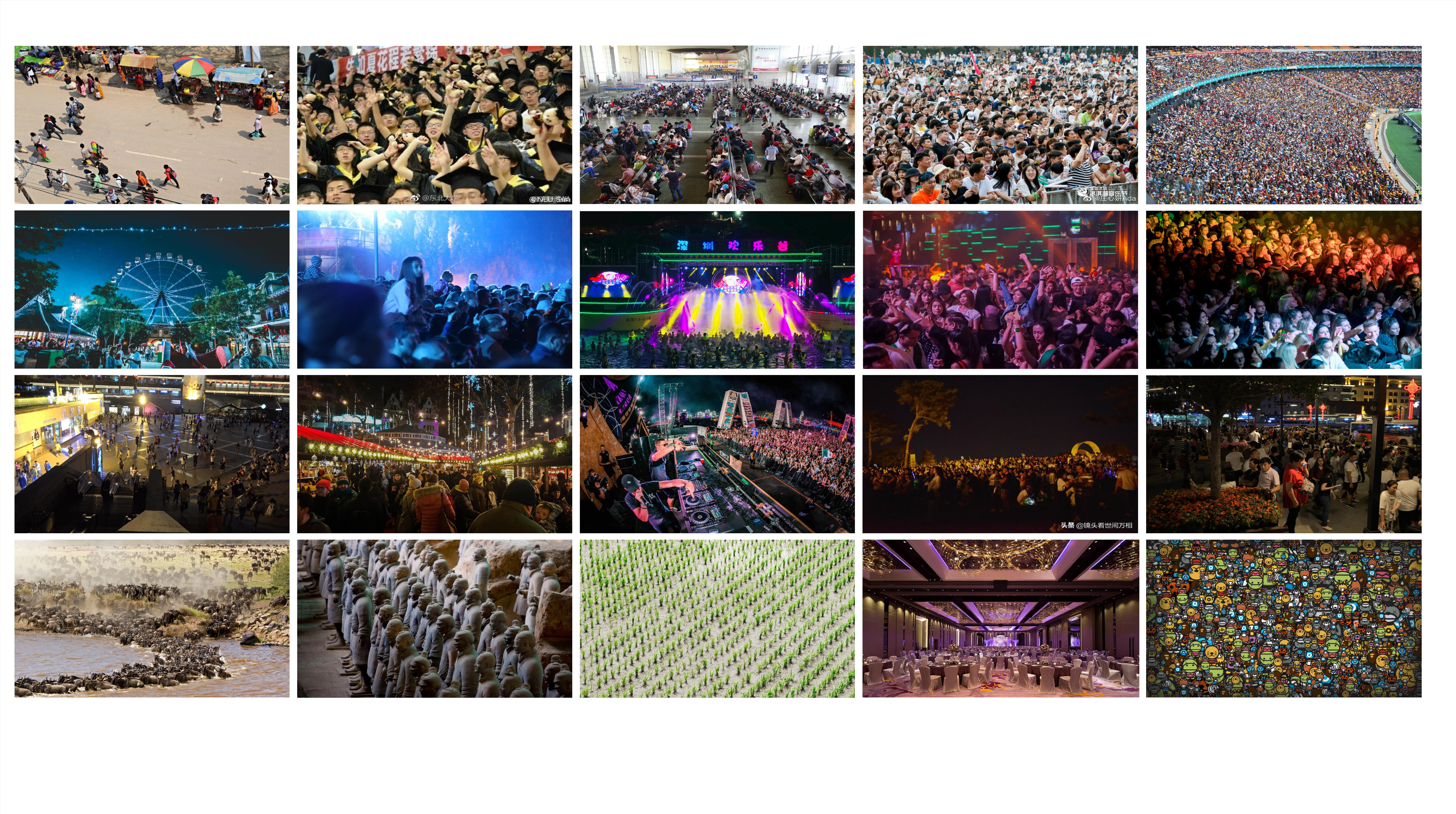}
	\caption{The display of the proposed NWPU-Crowd dataset. Column 1 shows some typical samples with normal lighting. The second and third column demonstrate the crowd scenes under the extreme brightness and low-luminance conditions, respectively. The last column illustrates the negative samples, including some scenes with densely arranged other objects. }\label{Fig-exemplars}
\end{figure*}

\subsection{General-scene Dataset} 

In addition to the above crowd images captured in specific scenes, there are also many general-scene crowd counting datasets, which are collected from the Internet. A remarkable aspect of general-scene is that the crowd density varies significantly, which ranges from $0$ to $20,000$. Besides, diversified scenarios, light and shadow conditions, and uneven crowd distribution in one single image are also distinctive attributes of these datasets.

The first general-scene dataset for crowd counting, UCF\_CC\_50~\cite{idrees2013multi}, is presented by Idrees \emph{et al.} in 2013. It only contains $50$ images, which is so small to train a robust deep learning model. Consequently, a larger crowd counting dataset becomes more significant nowadays. Zhang~\textit{et al.} propose ShanghaiTech Part A~\cite{zhang2016single}, which is constructed of $482$ images crawled from the Internet. Although its average number of labeled heads in each image is smaller than UCV\_CC\_50, it contains more pictures and larger number of labeled head points.
For further research on the extremely congested crowd counting, UCF-QNRF~\cite{idrees2018composition} is presented by Idrees~\textit{et al.} It is composed of $1,525$ images with more than $1,251,642$ label points. The average number of pedestrians per image is $815$, and the maximum number reaches $12,865$. Aiming at the small size of crowd images, Crowd Surveillance~\cite{yan2019perspective} build a large-scale dataset containing $13,945$ images, which provides regions of interest (ROI) for each image to keep out these blobs that are ambiguous for training or testing. In addition to the above datasets, Sindagi~\textit{et al.} introduce a new dataset for unconstrained crowd counting, JHU-CROWD, including $4,250$ samples. All images are annotated from the image and head level. For the former level, they label the scenario (\textit{mall, stadium}, etc.) and weather conditions. For the head level, the annotation information includes not only head locations but also occlusion, size, and blur attributes.

\section{NWPU-Crowd Dataset}
\label{NWPU}

This section describes the proposed NWPU-Crowd from four perspectives: data collection/specification, annotation tool, statistical analysis, data split and evaluation protocol. 

\subsection{Data Collection and Specification}

\textbf{Data Source.} Our data are collected from self-shooting and the Internet. For the former, $\sim\!2,000$ images and $\sim\!200$ video sequences are captured in some populous Chinese cities, including Beijing, Shanghai, Chongqing, Xi'an, and Zhengzhou, containing some typical crowd scenes, such as resort, walking street, campus, mall, plaza, museum, station. However, extremely congested crowd scenes are not the norm in real life, which is hard to capture via self-shooting. Therefore, we also collect $\sim\!8,000$ samples from some image search engines (Google, Baidu, Bing, Sougou, \emph{etc.}) via the typical query keywords related to the crowd. Table \ref{Table-search} lists the primary data source websites and the corresponding keywords. The third row in the table records some Chinese websites and keywords. Finally, by the above two methods, $10,257$ raw images are obtained. 

\begin{table}[htbp]
	
	\centering
	\caption{The query keywords on some typical search engines.}
	\scriptsize
	\setlength{\tabcolsep}{0.2mm}{
	\begin{tabular}{c|c|c}
		\whline
		\multirow{2}{*}{Data Source}	  &\multicolumn{2}{c}{Keywords} \\
		\cline{2-3}
		& Crowd & Negative Sample\\
		\whline
		\makecell[c]{google, baidu, \\ bing, pxhere, \\ pixabay...}  & \makecell[c]{crowd, congestion, hundreds/thousands of people, \\speech, conference,  ceremony, stadium, gathering, \\parade, demonstration, protest,   carnival,  beer festival, \\hajj, NBA, WorldCup, NFL, EPL, Super Bowl, \emph{etc.}} & \makecell[c]{dense, migration, \\ fish school,   \\empty scenes, \\ flowers, \emph{etc.}}\\
		\hline
		\makecell[c]{baidu, weibo, \\ sogou, so, \\ wallhere... }   & \begin{CJK}{UTF8}{gbsn}\makecell[c]{人群，拥挤，春运，军训，典礼，祭祀，庙会，\\游客，万人，千人，大赛，运动会，候车厅，\\音乐会/节，见面会，人从众，黄金周，招聘会，\\万人空巷，人山人海，摩肩接踵，水泄不通……}\end{CJK} &\begin{CJK}{UTF8}{gbsn}\makecell[c]{礼堂，动物迁徙，\\花海，动漫人物，\\餐厅，空无一人，\\密集排列……}\end{CJK}\\
		\whline
	\end{tabular}}
	\label{Table-search}
\end{table}

\noindent\textbf{Data Deduplication and Cleaning.} We employ four individuals to download data from the Internet on non-overlapping websites. Even so, there are still some images that contain the same content. Besides, some congested datasets (UCF\_CC\_50, Shanghai Tech Part A, and UCF-QNRF), are also crawled from the Internet, e.g., Flickr, Google, \emph{etc.} For avoiding the problem of data duplication, we perform an effective strategy to measure the similarity between two images, which is inspired by Perceptual Loss \cite{johnson2016perceptual}. Specifically, for each image, the layer-wise VGG-16 \cite{simonyan2014very} features (from conv1 to conv5\_3 layer) are extracted. Given two resized samples $i_x$ and $i_y$ with the resolution of $224 \times 224$, the similarity is defined as follows:

\begin{equation}
\scriptsize
\begin{array}{l}
\begin{aligned}
D\left( {{i_x},{i_y}} \right) = \sum\limits_{j \in L} {\frac{1}{{{C_j}{H_j}{W_j}}}} \left\| {{\psi _j}({i_x}) - {\psi _j}({i_y})} \right\|_2^2,
\end{aligned}\label{similarity}
\end{array}
\end{equation}
where $L$ is the set of the last activation layer in five groups of VGG-16 network, namely $L = \left\{ {{\psi _j}|j = 1,2,...,5} \right\} = \left\{ {relu1\_2,relu2\_2,relu3\_3,relu4\_3,relu5\_3} \right\}$. ${\psi _j}({i_x})$ and ${\psi _j}({i_y})$ denote layer ${\psi _j}$'s outputs (feature maps) for sample $i_x$ and $i_y$, respectively. ${C_j}$, ${H_j}$ and ${W_j}$ are the size of ${\psi _j}({i_x})$ at three axes: channel, height and width. If $D\left( {{i_x},{i_y}} \right)<5$, these two samples are considered to have similar contents. As a result, one of the two is removed from the dataset.

Then remove excess similar images by computing the distance of the feature between any two samples. Furthermore, some blurred images that are difficult to recognize the head location are also removed. Consequently, we obtain $5,109$ valid images.

\subsection{Data Annotation}
\textbf{Annotation tools:} For conveniently annotating head points in the crowd images, an online efficient annotation tool is developed based on HTML5 + Javascript + Python. This tool supports two types of label form, namely point and bounding box. During the annotation process, each image is flexibly zoomed in/out to annotate head with different scales, and it is divided into $16 \times 16$ small blocks at most, which allows annotators to label the head under five scales: $2^i$ (i=0,1,2,3,4) times size of the original image. It effectively prompts annotation speed and quality. The more detailed description is shown in the video demo of our provided supplementary material.

\noindent\textbf{Point-wise annotation:} The entire annotation process has two stages: labeling and refinement. Firstly, there are 30 annotators involved in the initial labeling process, which costs $2,100$ hours totally to annotate all collected images. After this, 6 individuals are employed to refine the preliminary annotations, which takes $150$ hours per refiner. In total, the entire annotation process costs $3,000$ human hours. 

\noindent\textbf{Box-level annotation and generation:} There are three steps to annotate box labels: 1) for each image, manually select $\sim 10\%$ typical points to draw their corresponding boxes, which can represent the scale variation in the whole scene; 2) for each point without box label, adopt a linear regression algorithm to obtain its box size based on its 8-nearest box-labeled neighbors; 3) manually refine the prediction box labels. Step 1) and 2) takes $1,000$ human hours in total.

Here, the step 2) is described as below: For a head point $P_0$ without box label, its 8-nearest box-labeled neighbors ($P_{1~8}$) are utilized to fit a linear regression algorithm \mbox{\cite{draper1998applied}}, in which the vertical axis coordinates are variable, and the box size is the dependent variable. According to the linear function and the vertical axis coordinate of $P_0$, the box size corresponding to $P_0$ can be obtained. We assume each box has a shape of a square, and the point coordinates are its center, and then the box can be obtained. Obviously, the linear regression is not reliable, so we should manually refine the predicted box labels again in Step 3). Then the linear regression and the manually refine will loop continuously until all boxes seem qualified. In the annotation stage, Step 2) and 3) are repeated four times.

\noindent\textbf{Discussion on annotation quality} In the field of crowd counting and localization, it is important how to ensure high-quality annotation, especially in some extremely congested scenes. In this work, we attempt to alleviate it from the two aspects: 1) the proposed tools support zooming in or out on an image with 1x~16x online. For congested region, the annotator can easily draw a box on a tiny or occluded object using zooming operation; 2) we conduct two stages of refinement in the point annotation, and repeat four times for linear estimation and refinement to minimize labeling errors in the box annotation.

\subsection{Data Characteristic}

\label{characteristic}

\begin{figure} 
	\centering 
	\subfigure[The distribution of counts on the four datasets.] { \label{Fig-hist-cnt} 
		\includegraphics[width=0.7\columnwidth]{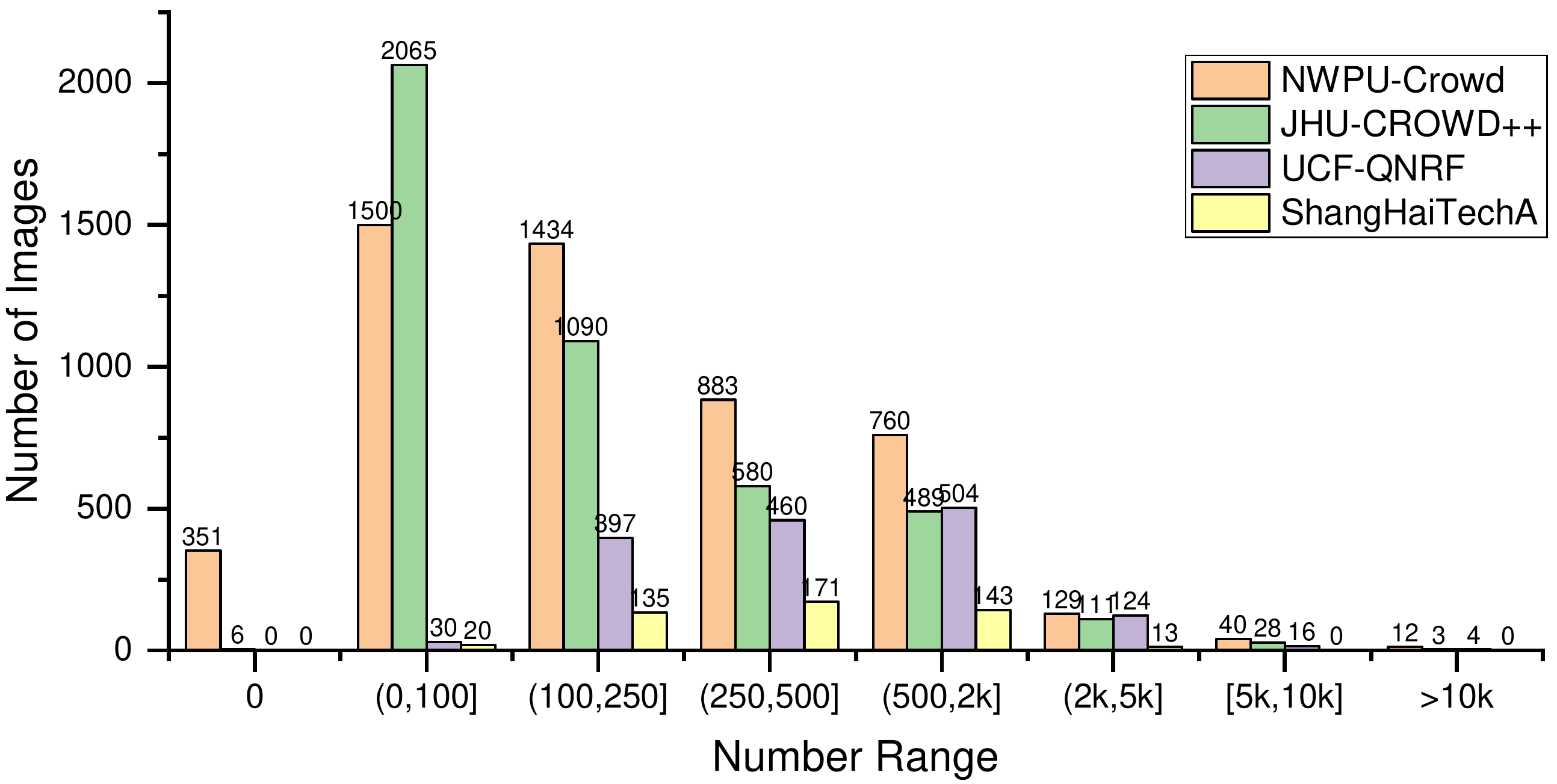}
	} 
	\subfigure[The distribution of the area (pixels) of head region. ] { \label{Fig-hist-box} 
		\includegraphics[width=0.7\columnwidth]{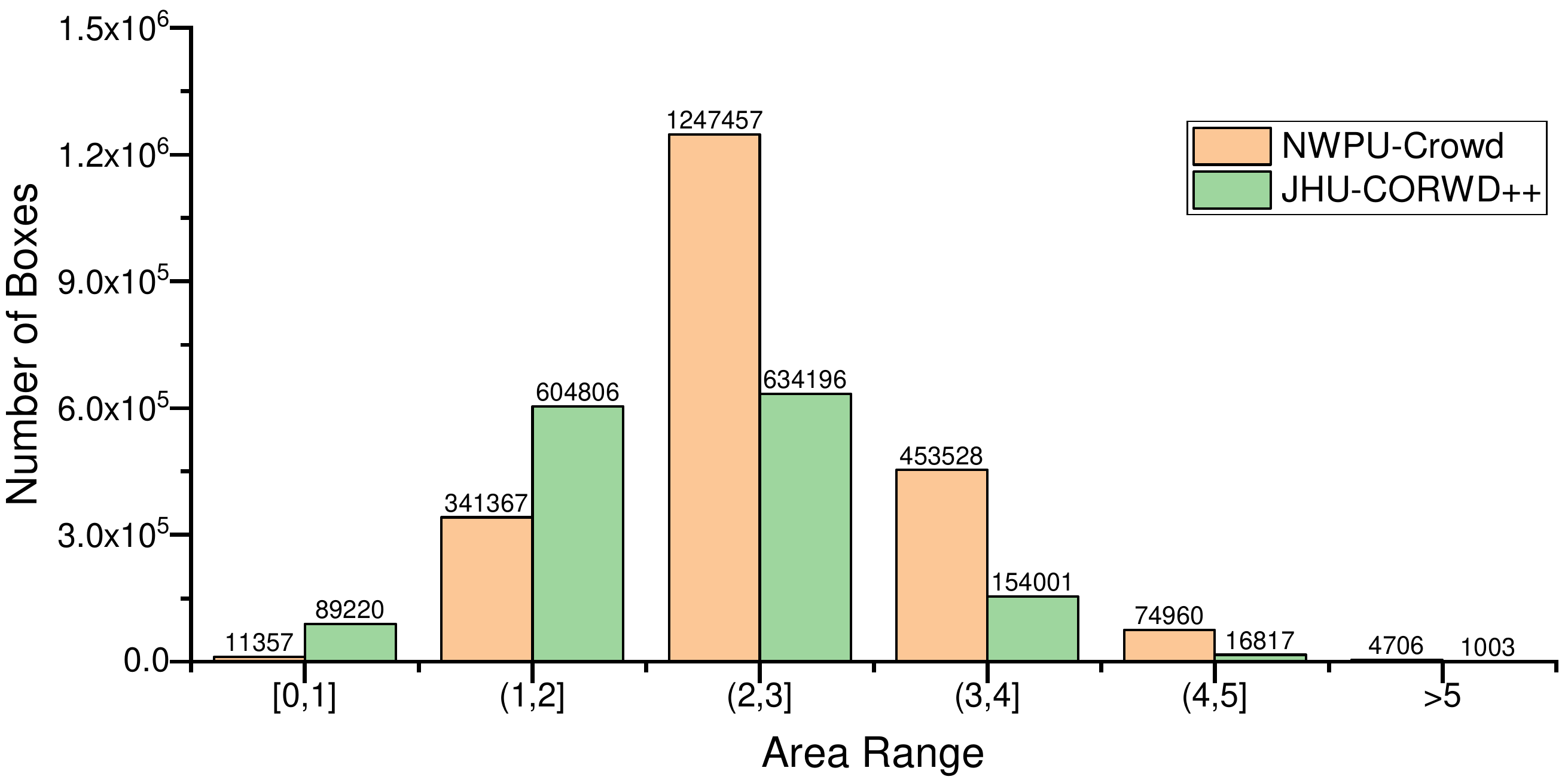}
	} 
	\caption{The statistical histogram of image-level counts and box-level area. The number in $x$ axis of Fig.(b) denotes $10^x$. For example, $[0,1]$ represents the range of box area is $[10^0,10^1]$.} 
	\label{Fig-sector} 
\end{figure}

NWPU-Crowd dataset consists of $5,109$ images, with $2,133,375$ annotated instances. Compared with the existing crowd counting datasets, it is the largest from the perspective of image and instance level. Fig. \ref{Fig-exemplars} respectively demonstrates four groups of typical samples from Row 1 to 4 in the dataset: normal-light, extreme-light, dark-light, and negative samples. Fig. \mbox{\ref{Fig-hist-cnt}} compares the number distribution of different counting range on four datasets: NWPU-Crowd, JHU-CROWD++ \mbox{\cite{sindagi2019pushing,sindagi2020jhu}}, UCF-QNRF \mbox{\cite{idrees2018composition}}, and ShanghaiTech Part A \mbox{\cite{zhang2016single}}. Except the bin of $(0,100]$, the number of images on NWPU-Crowd is much larger than that on the other three datasets. Fig. \mbox{\ref{Fig-hist-box}} shows the distributions of the box area in NWPU-Crowd and JHU-CROWD. From the orange bars, more than $50\%$ of  boxes areas are in the range of $(10^2,10^3]$ pixels. Since the average resolution of NWPU-Crowdis is higher than that of JHU-CROWD, the numbers of large-scale heads are more. The larger scale provides more detailed head-structure information, which will aid the model to achieve better performance. 

In addition to data volume and scale distribution, there are four more advantages in NWPU-Crowd:

\begin{enumerate}
	\item[1)] \textbf{Negative Samples.} NWPU-Crowd introduces $351$ negative samples (namely nobody scenes), which are similar to congested crowd scenes in terms of texture features. It effectively improves the generalization of counting models while applied in the real world. These samples contain animal migration, fake crowd scenes (sculpture, Terra-Cotta Warriors, 2-D cartoon figure, etc.), empty hall, and other scenes with densely arranged objects that are not the person. 
	
	\item[2)] \textbf{Fair Evaluation.} For a fair evaluation, the labels of the test set are not public. Therefore, we develop an online evaluation benchmark website that allows researchers to submit their estimation results of the test set. The benchmark can calculate the error between presented results and ground truth, and list them on a scoreboard.
	
	\item[3)] \textbf{Higher Resolution.} The proposed dataset collects high-quality and high-resolution scenes, which is entailed for extremely congested crowd counting. From Table \ref{data-com}, the average resolution of NWPU-Crowd is $2191 \times 3209$, which is larger than that of other datasets. Specifically, the maximum image size is $4028 \times 19044$.
	
	\item[4)] \textbf{Large Appearance Variation.} The number of people ranges from $0$ to $20,033$, which means large appearance variations within the data. Notably, the smallest head occupies only $4$ pixels, but the largest head covers $1.2\!\times\!10^7$ pixels. In the whole dataset, the ratio of the area  of the largest and smallest head in the same image is $3.8\!\times\!10^5$.

\end{enumerate}

In summary, NWPU-Crowd is one of the largest and most challenging crowd counting/localization datasets at present.

\subsection{Data Split and Evaluation Protocol}
NWPU-Crowd Dataset is randomly split into three parts, namely \emph{training}, \emph{validation} and \emph{test} sets, which respectively contain $3,109$, $500$ and $1,500$ images. To be specific, each image is randomly assigned to a specific set with the corresponding probability (followed by $0.6$, $0.1$ and $0.3$ for the three subsets) until the number reaches the upper bound. This strategy ensures that the statistics (such as data distribution, the average value of resolutions/counts) of the subset are almost the same. 

\noindent\textbf{Counting Metrics \,\,\,} Following some previous works, we adopt three metrics to evaluate the counting performance, which are Mean Absolute Error (MAE), Mean Squared Error (MSE), and mean Normalized Absolute Error (NAE). They can be formulated as follows: 

\begin{equation}
\tiny
\begin{array}{l}
\begin{aligned}
MAE = \frac{1}{N}\sum\limits_{i = 1}^N {\left| {{y_i} - {{\hat y}_i}} \right|}, MSE = \sqrt {\frac{1}{N}\sum\limits_{i = 1}^N {{{\left| {{y_i} - {{\hat y}_i}} \right|}^2}} }, NAE = \frac{1}{N}\sum\limits_{i = 1}^N {\frac{\left| {{y_i} - {{\hat y}_i}} \right|}{{y_i}}},
\end{aligned}\label{MAE}
\end{array}
\end{equation}
where $N$ is the number of images, ${{y_i}}$ is the counting label of people and ${{{\hat y}_i}}$ is the estimated value for the $i$-th test image. Since NWPU-Crowd contains quite a few negative samples, NAE's calculation does not contain them to avoid zero denominators. 

In addition to the aforementioned overall evaluation on \emph{the test set}, we further assess the model from different perspectives: scene level and luminance. The former have five classes according to the number of people: $0$, $(0,100]$, $(100,500]$,  $(500,5000]$, and more than $5000$. The latter have three classes based on luminance value in the YUV color space: $[0,0.25]$, $(0.25,0.5]$,and $(0.5,0.75]$. The two attribute labels are assigned to each image according to their annotated counting number and image contents. For each class in a specific perspective, MAE, MSE, and NAE are applied to the corresponding samples in \emph{the test set}. Take the luminance attribute as an example, the average values of MAE, MSE, and NAE at the three categories can reflect counting models' sensitivity to the luminance variation. Similar to the overall metrics, the negative samples are excluded during the calculation of NAE.

\noindent\textbf{Localization Metrics \,\,\,} For the crowd localization task, we adopt the box-level Precision, Recall and F1-measure to evaluate the localization performance. Given two point sets from prediction results $P_p$ and ground truth $P_g$, we firstly construct a Bipartite Graph ${G_{p,s}}$ for the two sets. Secondly, we compute the distance matrix of $P_p$ and $P_p$. If the distance between $p_p \in P_p$ and $p_g \in P_g$ is less than the predefined distance threshold $\sigma$, we think $p_p$ and $p_g$ are successfully matched. Corresponding to each element of the distance matrix, we obtain a boolean match matrix (True and False denote matched and non-matched). Finally, we can get a Maximum Bipartite Matching for ${G_{p,s}}$ by implementing the Hungarian algorithm \footnote{Note that Hungarian algorithm's matching result is not unique. However, the number of TP, FP and FN is the same for different matching results. Considering that for saving computation time, we perform Hungarian algorithm on match matrix instead of the distance matrix.} the match matrix and count the number of True Positive (TP), False Positive (FP) and False Negative (FN).  In our evaluation, for each head with the size of width $w$ and height $h$, we define two threshold $\sigma_s=min(w,h)/2$ and $\sigma_l={\sqrt {{w^2} + {h^2}}}/2 $. The former is a stricter criterion than the latter.

Similar to the category-wise counting evaluation at the image level, we propose a scale-sensitive evaluation scheme at the box level for the localization task. To be specific, all heads are divided into six categories according to their corresponding box areas: $[10^0,10^1]$, $(10^1,10^2]$,  $(10^2,10^3]$,  $(10^3,10^4]$,  $(10^4,10^5]$, and more than $10^5$. For each category, the Recall is calculated separately.

Different from the previous localization metrics \mbox{\cite{idrees2018composition,liu2019recurrent}}, the $\sigma$ in this paper is adaptive, which is defined by the real head area. In addition, the performance on different scale classes are reported, which helps researchers analyze the model more deeply. In summary, our evaluation is more reasonable than the traditional methods.

\section{Experiments on Counting}
\label{exp}

In this section, we train ten mainstream open-sourced methods on the proposed NWPU-Crowd and submit their results on the evaluation benchmark. Besides, the further experimental analysis and visualization results on \emph{the validation set} are discussed.  

\subsection{Mainstream Methods Involved in Evaluation}

\begin{figure*}
	\centering
	\includegraphics[width=0.9\textwidth]{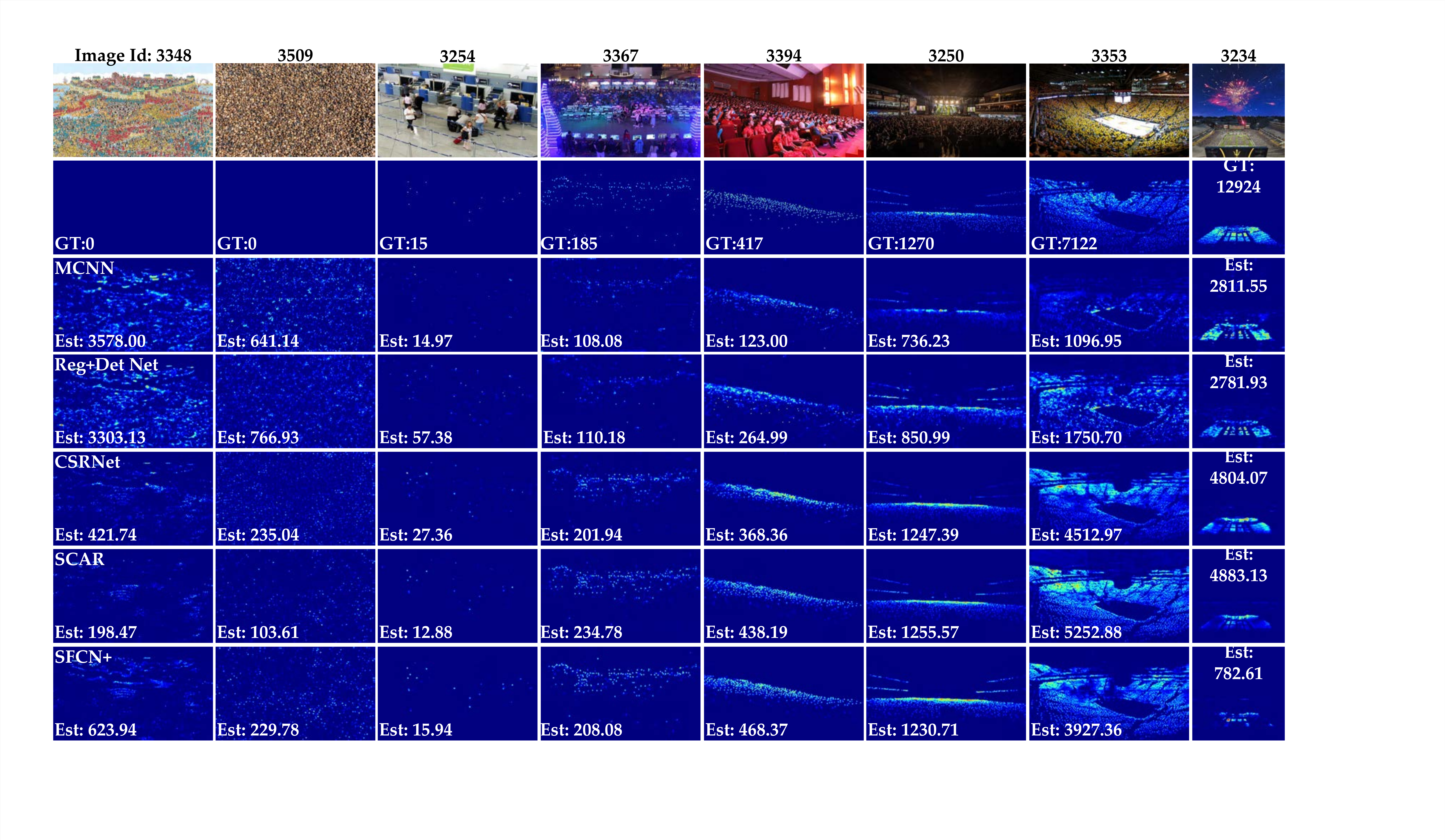}
	\caption{The eight groups of visualization results of some selected methods on \emph{the validation set}. }\label{Fig-results}
\end{figure*}

\textbf{MCNN \cite{zhang2016single}:} Multi-Column Convolutional Neural Network. It is a classical and lightweight counting model, proposed by Zhang \emph{et al.} in 2016. Different from the original MCNN, the RGB images are fed into the network.

\noindent\textbf{SANet \cite{cao2018scale}:} Scale Aggregation Network. SANet is an efficient encoder-decoder network with Instance Normalization for crowd counting, which combines the MSE loss and SSIM loss to output the high-quality density map. 

\noindent\textbf{PCC Net \cite{gao2019pcc}:} Perspective Crowd Counting Network. It is a multi-task network, which tackles the following tasks: density-level classification, head region segmentation, and density map regression. The authors provide two versions, a lightweight from scratch and VGG-16 backbone. 

\noindent\textbf{Reg+Det Net \mbox{\cite{liu2018decidenet}}:} a subnet of DecideNet. It consists of two branches: Regression and Detection Network. The former is a light-weight network for density estimation, and the latter focuses on head detection via on Faster R-CNN (ResNet-101) \mbox{\cite{ren2015faster}}.

\noindent\textbf{C3F-VGG \cite{gao2019c}:} A simple baseline based on VGG-16 backbone for crowd counting. C3F-VGG consists of the first 10 layers of VGG-16 \cite{simonyan2014very} as image feature extractor and two convolutional layers with a kernel size of 1 for regressing the density map.

\noindent\textbf{CSRNet \cite{li2018csrnet}:} Congested Scene Recognition Network. CSRNet is a classical and efficient crowd counter, proposed by Li \emph{et al.} in 2016. The authors design a Dilatation Module and add it to the top of the VGG-16 backbone. This network significantly improves performance in the field of crowd counting. 

\noindent\textbf{CANNet \cite{liu2019context}:} Context-Aware Network. CANNet combines the features of multiple streams using different respective field sizes. It encodes the multi-scale contextual information of the crowd scenes and yields a new record on the mainstream datasets. 

\noindent\textbf{SCAR \cite{gao2019scar}: } Spatial-/Channel-wise Attention Regression Networks. SCAR utilizes the self-attention module \cite{wang2018non} on the spatial and channel axis to encode the large-range contextual information. The well-designed attention models effectively extracts discriminative features and alleviates mistaken estimations.

\noindent\textbf{BL \cite{ma2019bayesian}:} Bayesian Loss for Crowd Count Estimation. Different from the traditional strategy for the generation of ground truth, BL design a loss function to directly using head point supervision. It achieves state-of-the-art performance on the UCF-QNRF dataset. 

\noindent\textbf{SFCN\dag\, \cite{wang2019learning}} Spatial Fully Convolutional Network with ResNet-101 \cite{he2016deep}. SFCN\dag \, is the only crowd counting model that uses ResNet-101 as a backbone, which shows the powerful capacity of density regression on the congested crowd scenes.

\subsection{Implementation Details}

In the experiments, for PCC Net \footnote{https://github.com/gjy3035/PCC-Net} and BL \footnote{https://github.com/ZhihengCV/Bayesian-Crowd-Counting}, the models are trained using the official codes and the default parameters. For SANet, we implement the $C^{3}$ Framework \cite{gao2019c} and follow the corresponding parameters to train them on NWPU-Crowd dataset. For DetNet, we train a head detector using this code \footnote{https://github.com/ruotianluo/pytorch-faster-rcnn}.

For other models, namely MCNN, RegNet, CSRNet, C3F-VGG, CANNet, SCAR, and SFCN\dag, they are reproduced in our counting experiments, which is developed based on $C^{3}$ Framework \cite{gao2019c}, an open-sourced crowd counting project using PyTorch \cite{paszke2019pytorch}. In the data pre-processing stage, the high-resolution images are resized to the 2048-px scale with the original aspect ratio. The density map is generated by a Gaussian kernel with a fixed size of 15 and the $\sigma$ of 4. For augmenting the data, during the training process, all images are randomly cropped with the size of $576\times768$, flipped horizontally, transformed to gray-scale images, and gamma corrected with a random value in $[0.4,2]$. To optimize the above counting networks, Adam algorithm \cite{kingma2014adam} is employed. Other parameters (such as learning rate, batch size) are reported in \url{https://github.com/gjy3035/NWPU-Crowd-Sample-Code}.

\subsection{Results Analysis on \emph{the Validation Set}}

\label{exp-val}

\textbf{Quantitative Results.} Here, we list the counting performance and density quality of all participation methods in Table \ref{Table-val}. For evaluating the quality of the density map, two popular criteria are adopted, Peak Signal-to-Noise Ratio (PSNR) and Structural Similarity in Image (SSIM) \cite{wang2004image}. Since BL \cite{ma2019bayesian} is supervised by point locations instead of density maps, PSNR and SSIM are not reported. In the calculation of PSNR, the negative samples are excluded to avoid zero denominators.

\begin{table}[htbp]
	\centering
	\footnotesize
	
	\caption{The performance of different models on \emph{the val set}.}
	
	\begin{tabular}{cIc|c|c|c}
		\whline
		\multirow{2}{*}{Method}	 &\multicolumn{4}{c}{\emph{the validation set}} \\
		\cline{2-5}
		&  MAE &MSE &PSNR &SSIM  \\
		\whline
		MCNN   &218.53 &700.61 &28.558 &0.875 \\
		\hline
		SANet  &171.16 &471.51  &29.228 &0.886	  \\
		\hline
		Reg+Det Net &245.8 &700.3  &28.862 &0.751 \\
		\hline	
		PCC-Net-light &141.37 &630.72 &29.745 &0.937 \\
		\whline
		C3F-VGG &105.79 &504.39 &29.977& 0.918  \\
		\hline
		CSRNet &104.89 &433.48 &29.901 &0.883  \\
		\hline
		PCC-Net-VGG&100.77 &573.19 &30.565 &0.941  \\
		\hline
		CANNet &93.58 &489.90 &30.428 &0.870  \\
		\hline
		SCAR &\textbf{81.57} &\textbf{397.92} &30.356 &0.920  \\
		\hline
		BL &93.64 &470.38 &-&- \\
		\whline
		SFCN\dag &95.46 &608.32 &\textbf{30.591} &\textbf{0.952}  \\			
		\whline
	\end{tabular}\label{Table-val}
\end{table}

From the table, we find SCAR \cite{gao2019scar} attains the best counting performance, MAE of 81.57 and MSE of 397.92. SFCN\dag \, \cite{wang2019learning} produces the most high-quality density maps, PSNR of 30.591 and SSIM of 0.952. For the three light models (MCNN, SANet, PCC-Net-light), we find that the last achieves the best SSIM (0.937), which even surpasses the SSIMs of some other VGG-based algorithms, such as C3F-VGG, CSRNet, CANNet, and SCAR. Similarly, PCC-Net-VGG is the best SSIM in the VGG-backbone methods. 

\noindent\textbf{Visualization Results.} Fig. \ref{Fig-results} demonstrates some predicted density maps of the eight methods. The first two columns are negative samples, and others are crowd scenes with different density levels. From the first two columns, almost all models perform poorly for negative samples, especially densely arranged objects. For humans, we can easily recognize that the two samples are mural and stones. But for the counting models, they cannot understand them. For the third column, although the predictions of these methods are good, there are still many mistaken errors in background regions. For the last two images that are extremely congested scenes, the estimation counts are far from the ground truth. SCAR is the most accurate method on \emph{the validation set}, but it is about $1,900$ and $8,000$ people away from the labels, respectively. For the extreme-luminance scenes (Image 3367, 3250, and 3353), there are quite a few estimation errors in the high-light or dark-light regions. In general, the ability of the current models to cope with the above hard samples needs to be further improved. 

\subsection{Leaderboard}

\label{exp-test}

\begin{table*}[htbp]
	\scriptsize
	\centering
	\caption{The leaderboard of the counting performance on the NWPU-Crowd \emph{test set}. In the ranking strategy, the Overall MAE is the primary key. ``FS'' represents that the model is trained From Scratch. $S0\!\sim\!S4$ respectively indicates five categories according to the different number range: $0$, $(0,100]$, ..., $\geq5000$. $L0\!\sim\!L2$ respectively denotes three luminance levels on \emph{the test set}: $[0,0.25]$, $(0.25,0.5]$, and $(0.5,0.75]$. Limited by the paper length, only MAE are reported in the category-wise results. The speed and FLOPs are computed on the input size of $576 \times 768$. The \textbf{bold} and  \underline{underline} fonts respectively represent the \textbf{first} and \underline{second} place.}
	\setlength{\tabcolsep}{1.2mm}{
		\begin{tabular}{c|cIc|c|cIc|cIc|cIc|c|c}
			\whline
			\multirow{2}{*}{Method}	&\multirow{2}{*}{Backbone} &\multicolumn{3}{cI}{Overall} &\multicolumn{2}{cI}{Scene Level (only MAE)} &\multicolumn{2}{cI}{Luminance (only MAE)} &\multirow{2}{*}{\makecell[c]{Model\\Size (M)}} &\multirow{2}{*}{\makecell[c]{Speed \\ (fps)}} &\multirow{2}{*}{GFLOPs} \\
			\cline{3-9}
			& & \textbf{MAE} &MSE &NAE  &Avg. & $S0 \sim S4$ &Avg.   &$L0 \sim L2$ &&  \\
			\whline
			MCNN  &FS &232.5 & 714.6 & 1.063 & 1171.9 & 356.0/72.1/103.5/509.5/4818.2 & 220.9 & 472.9/230.1/181.6 &\textbf{0.133} &\textbf{129.0} & \textbf{11.867} \\
			\hline
			SANet  &FS &190.6 & 491.4 & 0.991 & 716.3 & 432.0/65.0/104.2/385.1/2595.4 & 153.8 & 254.2/192.3/169.7 & {1.389} & 10.8 & \underline{40.195}\\
			\hline
			Reg+Det Net  &Hybrid &264.9  &759.0  &1.770  &1242.5  &443.0/125.5/140.5/461.5/5036.6  &313.6  &464.2/267.4/209.1  & 189.6 &4.2 & 263.079 \\
			\hline
			PCC-Net-light  &FS & 167.4 & 566.2 & 0.444 & 944.9 & 85.3/25.6/80.4/424.2/4108.9 & 141.2 & 253.1/167.9/144.9 & \underline{0.504}&12.6 &{72.797} \\
			\whline
			C3F-VGG  &VGG-16 & 127.0 & 439.6 & 0.411 & {666.9} & 140.9/26.5/58.0/307.1/2801.8 & 127.9 & 296.1/125.3/91.3 & 7.701 &\underline{47.2} & 123.524\\
			\hline	
			CSRNet  &VGG-16 &121.3 & \underline{387.8} & 0.604 & \textbf{522.7} & 176.0/35.8/59.8/285.8/2055.8 & 112.0 & 232.4/121.0/95.5 &16.263 &26.1 & 182.695 \\
			\hline
			PCC-Net-VGG  &VGG-16 & 112.3 & 457.0 & \underline{0.251} & 777.6 & 103.9/13.7/42.0/259.5/3469.1 & 111.0 & 251.3/111.0/82.6 & 10.207 &24.0 & 145.157 \\				
			\hline
			CANNet  &VGG-16 & {106.3} & \textbf{386.5} & 0.295 &\underline{612.2} & 82.6/14.7/46.6/269.7/2647.0 &\textbf{102.1} & 222.1/104.9/82.3 &18.103 &22.0 & 193.580 \\
			\hline
			SCAR  &VGG-16 &110.0 & 495.3 & 0.288 & 718.3 & 122.9/16.7/46.0/241.7/3164.3 & \underline{102.3} & 223.7/112.7/73.9 &16.287 &24.5 & 182.856  \\
			\hline
			BL  &VGG-19 &\textbf{105.4}  &454.2  &\textbf{0.203} & 750.5 & 66.5/8.7/41.2/249.9/3386.4 & 115.8 & 293.4/102.7/68.0
			&21.449 & {34.7} & 182.186 \\
			\whline
			SFCN\dag  &ResNet-101 & \underline{105.7} & {424.1} & {0.254} & 712.7 & 54.2/14.8/44.4/249.6/3200.5 & {106.8} & 245.9/103.4/78.8 &38.597 &8.8 & 272.763 \\
			\whline
		\end{tabular}
		
	}
	\label{Table-lb}
\end{table*}

Table \mbox{\ref{Table-lb}} reports the results of five methods on \emph{the test set}. It lists the overall performance (MAE, MSE, and NAE), category-wise MAE on the attribute of scene level and luminance, model size, speed (inference time) and floating-point operations per second (FLOPs) \footnote{For PCC-Net, we remove the useless layers (classification and segmentation modules) to compute the last three items: model size, speed and FLOPs.}. Compared with the results of \emph{the validation set}, we find that the ordering has changed significantly. Although SCAR attains the best results of MAE and MSE on \emph{the validation set}, the performance on \emph{the test set} is not good. For the primary key (overall MAE), BL, SFCN\dag\, and CANNet occupy the top three on \emph{the test set}. 

From the category-wise results of Scene Level, we find that all methods perform poorly in S0 (negative samples), S3 ($(500,5000]$) and S4 ($\geq5000$), which causes that the average value of category-wise MAE is larger than the overall MAE (SFCN\dag: $712.7$ \emph{v.s.} $105.7$). Besides, this phenomenon shows that negative samples and congested scenes are more challenging than sparse crowd images. Similarly, for the luminance classes, the MAE of $L0$ ($[0,0.25]$) is larger than that of $L1$ and $L2$. In other words, the counters work better under the standard luminance than under the low-luminance scenes. More detailed results are shown in \url{https://www.crowdbenchmark.com/nwpucrowd.html}.

\subsection{Performance Impact between Different Scenes}

\label{inter}

From Section \ref{exp-val} and \ref{exp-test}, we find that two interesting phenomena worth attention: 1) Negative samples are prone to be mistakenly estimated; 2) The data with different scene attributes (namely density level) significantly affect each other. In this section, we conduct two experiments using a simple baseline model, C3F-VGG \cite{gao2019c}, to explore the above problems.

\noindent\textbf{Phenomenon 1. } For the firth problem, the main reason is that the negative samples contain densely arranged objects, which is similar to the congested crowd scenes. As we all know, most existing counting models focus on texture information and local patterns for congested regions. To verify our thoughts, we design three groups of experiments to explore which samples affect the performance of negative samples. To be specific, we train three C3F-VGG counters on different combination training data: $S0 + S1$, $S0 + S2$, and $S0 + S3 + S4$ (considering that the number of S4 is small, so we integrate S3 and S4). Then the evaluation is performed on \emph{the validation set}. Finally, the corresponding performance is listed in Table \ref{Table-combintion}. From it, the MAE on Negative Sample ($S0$) increases from $18.54$ to $147.53$ as the density of positive samples increases.

\begin{table}[htbp]
	\centering
	\footnotesize
	
	\caption{The MAE of the different training data on \emph{the val set}.}
	\setlength{\tabcolsep}{1.5mm}{
	\begin{tabular}{cIc|c|c|c|c}
		\whline
		\multirow{2}{*}{Combination}	 &\multicolumn{5}{c}{\emph{the validation set}} \\
		\cline{2-6}
		&  $S0$ & $S1$ &$S2$ &$S3$ &$S4$ \\
		\whline
		$S0+...+S4$ &61.25 &\underline{28.53} &\underline{51.91} &\underline{188.66} &\underline{3730.56} \\ 
		$S1+...+S4$ &- &33.12 &68.63 &238.88 &3997.57 \\    
		   
		\whline
		$S0 + S1$   &18.54 &\underline{16.44} &-& -&- \\
		$S1$ &- &21.00 &- &- & -\\
		\whline
		$S0 + S2$  &64.68 &- &\underline{49.97} &- & - \\
		$S2$ &- &- &51.01 &- & -\\
		\whline	
		$S0 + S3 + S4$ &147.53 &- &- &\underline{174.49} &\underline{2882.23}\\	
		$S3+S4$ &- &- &- &185.87 & 3488.25\\
		\whline
		
	\end{tabular}}\label{Table-combintion}
\end{table}

\noindent\textbf{Phenomenon 2. } For the second issue, we train the counting models only using the data with a single category, $S1$, $S2$, and $S3 + S4$ respectively. Removing the impacts of the negative samples, the model is trained on the data of $S1 + S2 + S3 + S4$. The concrete performance is illustrated in Table \ref{Table-combintion}. According to the results, training each class individually is far better than training together. To be specific, MAE decreases by \textbf{36.6\%}, \textbf{25.7\%}, \textbf{22.2\%} and \textbf{12.7\%} on the four classes, respectively. The main reason is that NWPU-Crowd contains more diverse crowd scenes than the previous datasets. There are large appearance variations in the dataset, especially the scales of the head. At present, the existing models can not tackle this problem well.

\subsection{The Effectiveness of Negative Samples} 

In Section \ref{characteristic}, we mention that the Negative Samples (``NS'' for short) can effectively improve the generalization ability of the model. Here, we conduct four groups of comparative experiments using C3F-VGG \cite{gao2019c} to verify this opinion. To be specific, there are four types of training data: $S1$, $S2$, $S3+S4$ and $S1+...+S4$. We respectively train the models for them using NS and without NS. In other words, we add $S0$ to the above four types of training data. The concrete results are reported in Table \ref{Table-combintion}. After introducing NS, the category-wise MAEs are significantly reduced. Take the last six rows as the examples, the MAE is respectively decreased by \textbf{21.7\%}, \textbf{2.0\%}, \textbf{6.1\%} and \textbf{17.4\%} on the category-wise evaluation. The main reason is that NS contains diverse background objects with different structured information, which can prompt the counting models to learn more discriminative features than ever before.

\subsection{Impact of Data Volume on Performance}

\begin{figure}[t]
	\centering
	\includegraphics[width=0.35\textwidth]{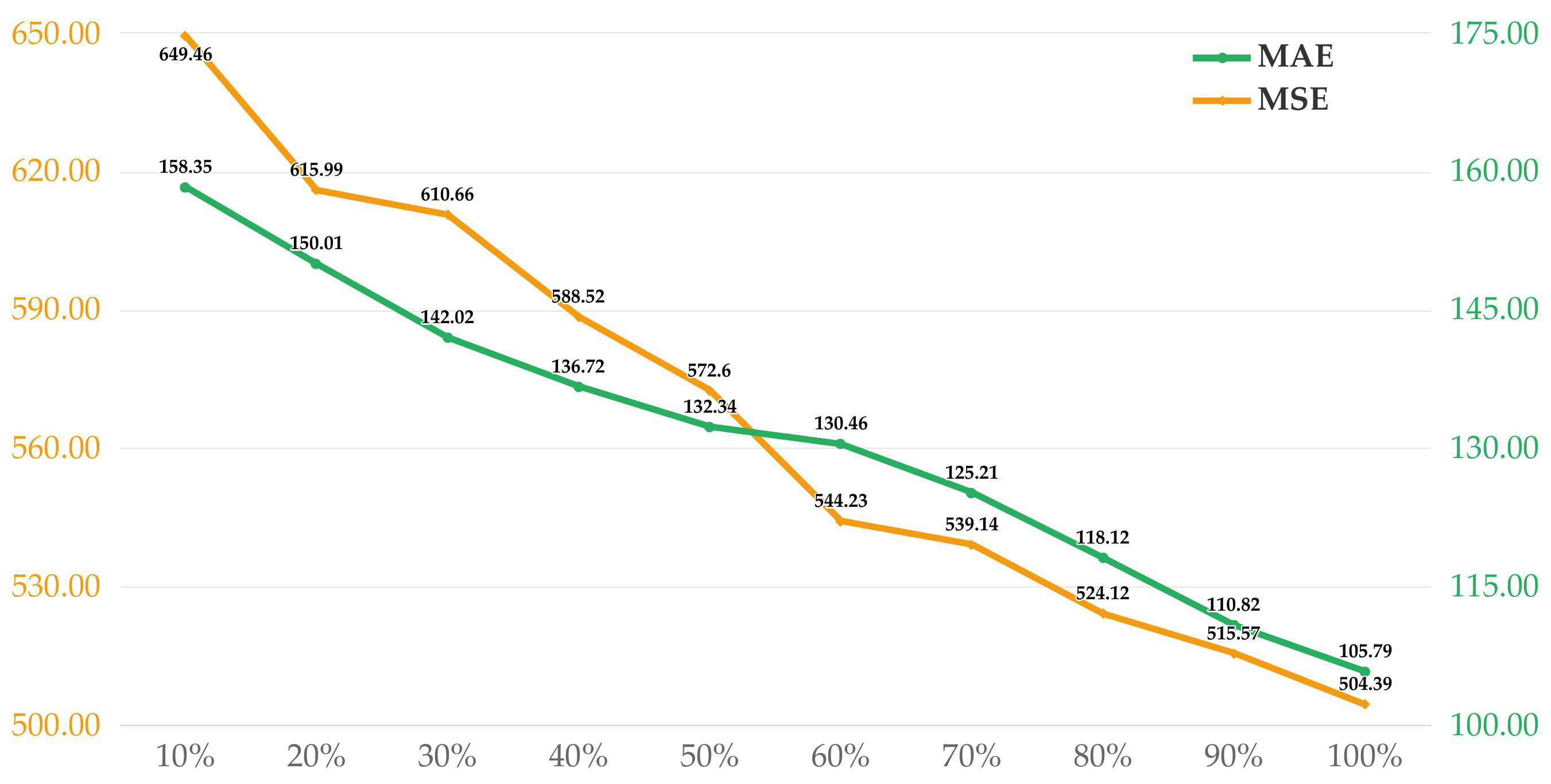}
	\caption{The results under different volumes of the training data on \emph{the validation set}. }\label{Fig-curve}
\end{figure}

Generally speaking, large amounts of diverse training data will prompt the model to learn more robust features, and then perform better in the wild. This is also our original intention to build a large-scale crowd counting and localization dataset. In this section, we explore the impact of different data volumes on counting performance. To be specific, we train ten C3F-VGG models using $10\%$, $20\%$, $30\%$, ..., $100\%$ training data, respectively. Then evaluate them on the \emph{validation set}. The performances (MAE and MSE) are demonstrated in Fig. \mbox{\ref{Fig-curve}}. From the figure, with the gradual increase of training data, the errors on the validation set also gradually decrease overall. By comparing the MAEs when using the $10\%$ and $100\%$ data, the error is significantly reduced from 158.35 to 105.79 (relative decrease of 33.2\%). Therefore, a large-scale dataset is very necessary for the community.

\section{Experiments on Localization}

Considering that the box-level labels are provided, we evaluate four crowd localization methods in this section. What's more, we analyze the quantitative and qualitative results of them. 

\subsection{Methods and Implementation Details}
\noindent\textbf{Faster RCNN \mbox{\cite{ren2015faster}}:} a general object detection framework, based on ResNet-101. It directly detects the head boxes, of which center is the prediction head location. We follow the original training parameters of this code \footnote{https://github.com/ruotianluo/pytorch-faster-rcnn}. In the forward process, the thresholds of confidence and nms are set as $0.8$ and $0.3$, respectively. 

\noindent\textbf{TinyFaces \mbox{\cite{8099649}}:} a tiny object detection framework, which focuses on tiny faces detection. We implement the third-party code \footnote{https://github.com/varunagrawal/tiny-faces-pytorch} to train a detector using the default parameters. The thresholds of confidence and nms are set as $0.8$ and $0.3$, respectively. 

\noindent\textbf{VGG+GPR \mbox{\cite{gao2019domain}}:} a two-stage method that consists of density map regression and point reconstruction based on Gaussian-kernel priors. C3F-VGG's \mbox{\cite{gao2019c}} training is the same as Section \mbox{\ref{exp}} and GPR using standard Gaussian kernel with a size of $15$.

\noindent\textbf{RAZ\_Loc \mbox{\cite{liu2019recurrent}}:} the localization branch of RAZNet, which consists of localization map classification and point post-processing based on finding high-confidence peaks. The training details follows RAZNet, and classification threshold is set as $0.5$.

\subsection{Results Analysis on \emph{the Validation Set}}

Table \mbox{\ref{Table-val-loc}} lists the localization and counting performance of four methods on the \emph{validation set}. For each head, its $\sigma_s$ is less than $\sigma_l$, which means that the former is more strict than the latter. Thus, the localization results of $\sigma_s$ are worse than that of $\sigma_l$. From the table, we find that the Precision of Faster RCNN is better than others, but they miss quite a few objects. RAZ\_Loc produces the best localization result, but its counting error is far from VGG+GPR. Detection-based methods' counting performance is the poorest in all plans.

\begin{table}[htbp]
	\centering
	\footnotesize
	
	\caption{The performance on \emph{the val set}. F1-m, Pre, Rec are short for F1-measure, Precision and Recall, respectively.}
	
	\begin{tabular}{cIc|c}
		\whline
		\multirow{2}{*}{Method}	 &localization & counting \\
		&  F1-m/Pre/Rec (\%) &MAE/MSE/NAE   \\
		\whline
		\multirow{2}{*}{Faster RCNN}   &$\sigma_l$:7.3/96.4/3.8  & \multirow{2}{*}{377.3/1051.2/0.798} \\
		\cline{2-2}
		&$\sigma_s$:6.8/90.0/3.5  & \\
		\hline
		\multirow{2}{*}{TinyFaces}   &$\sigma_l$: 59.8/54.3/66.6  & \multirow{2}{*}{240.4/736.2/0.962} \\
		\cline{2-2}
		&$\sigma_s$: 55.3/50.2/61.7 & \\
		\hline
		\multirow{2}{*}{VGG+GPR}   &$\sigma_l$: 56.3/61.0/52.2  & \multirow{2}{*}{105.8/504.4/0.931} \\
		\cline{2-2}
		&$\sigma_s$: 46.0/49.9/42.7 & \\
		\hline
		\multirow{2}{*}{RAZ\_Loc}   &$\sigma_l$: 62.5/69.2/56.9  & \multirow{2}{*}{128.7/665.4/0.508} \\
		\cline{2-2}
		&$\sigma_s$: 54.5/60.5/49.6 & \\
		\whline

	\end{tabular}\label{Table-val-loc}
\end{table}

\begin{figure} [htbp]
	\centering
	\includegraphics[width=0.48\textwidth]{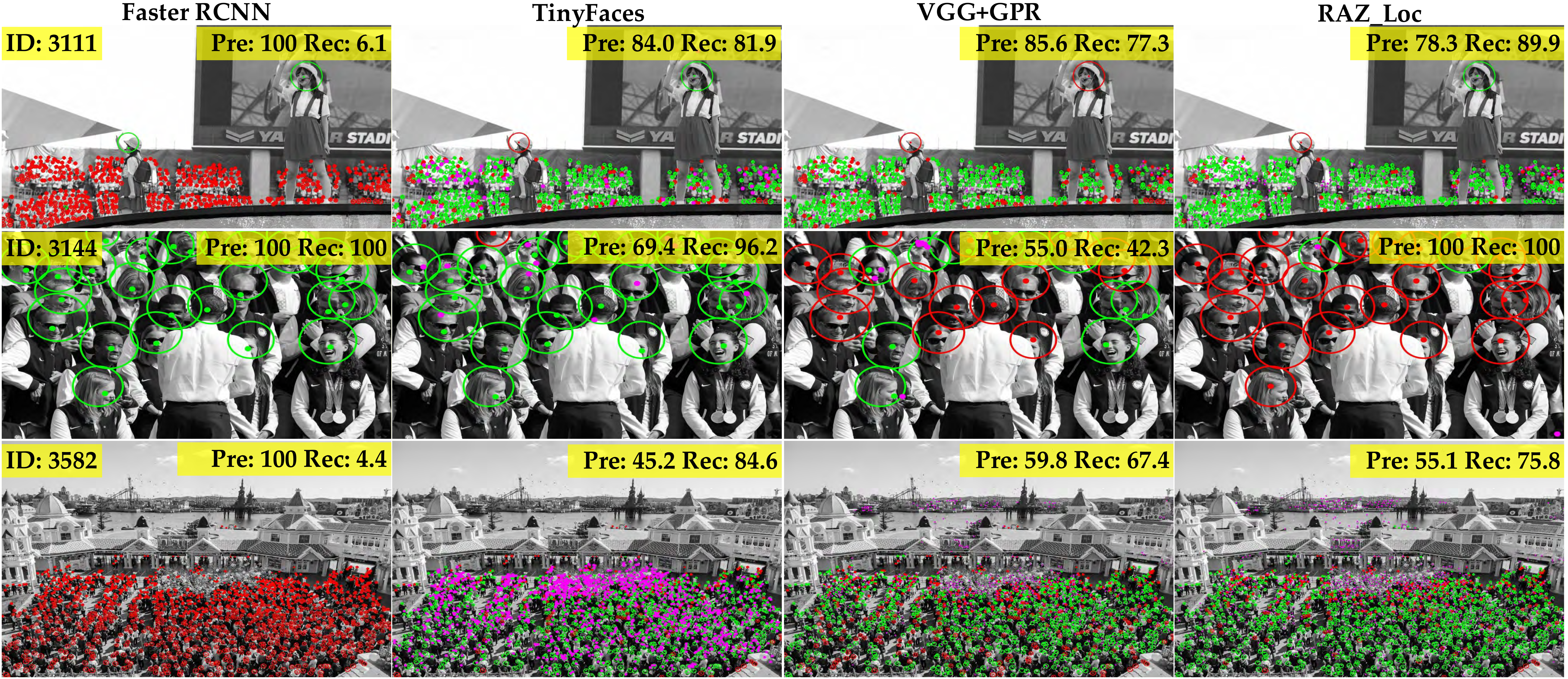}
	\caption{The three groups of qualitative localization results on \emph{the validation set}. The Green point is true positive, which is inside the green circle (its center is the groundtruth position and its radius is $\sigma_l$); the red points and the corresponding circles are false negative; the magenta points are false positive. (For a better comparison, we transform RGB-color images to gray-scale images.)}\label{Fig-vis-loc}
\end{figure}

\begin{table*}[htbp]
	\scriptsize
	\centering
	\caption{The leaderboard of the localization performance on the NWPU-Crowd test set. In the ranking strategy, the Overall F1-measure is the primary key under $\sigma_l$, which is bold font in Row 2 of the table.  $A0 \sim A4$ respectively indicates six categories according to the different head area ranges: $[10^0,10^1]$, $(10^1,10^2]$,  $(10^2,10^3]$,  $(10^3,10^4]$,  $(10^4,10^5]$, and $>10^5$. More detailed results will be reported in \url{https://www.crowdbenchmark.com/nwpucrowdloc.html}, which is under deployment.  \\ \textbf{*}: Since the counts is transformed to integer data, the performance is slightly different from Table \ref{Table-lb}.   }
	\begin{tabular}{cIc|c|cIc|cIc|c}
		\whline
		\multirow{2}{*}{Method}	&\multirow{2}{*}{Backbone} &\multicolumn{2}{cI}{Training Labels} &\multicolumn{2}{cI}{Overall ($\sigma_l$)}  &\multicolumn{2}{c}{Box Level (only Rec under $\sigma_l$) (\%)}  \\ 
		\cline{3-8}
		& &Point &Box & \textbf{F1-m}/Pre/Rec (\%)  &MAE/MSE/NAE & Avg. &$A0 \sim A5$\\
		\whline
		Faster RCNN  &ResNet-101 &\xmark  & \rmark & 6.7/\textbf{95.8}/3.5 &414.2/1063.7/0.791 &18.2 &0/0.002/0.4/7.9/37.2/63.5  \\
		\hline
		TinyFaces  &ResNet-101  & \xmark &  \rmark &56.7/52.9/\textbf{61.1}  &272.4/764.9/0.750  & \textbf{59.8} & 4.2/22.6/\textbf{59.1}/\textbf{90.0}/\textbf{93.1}/\textbf{89.6}    \\
		\hline
		VGG+GPR  &VGG-16 &\rmark  & \xmark & 52.5/55.8/49.6  &\textbf{127.3}/\textbf{439.9}/0.410*  & 37.4 & 3.1/27.2/49.1/68.7/49.8/26.3 \\
		\hline
		RAZ\_Loc  &VGG-16 &\rmark  & \xmark &\textbf{59.8}/66.6/54.3 &151.5/634.7/\textbf{0.305} &42.4 & \textbf{5.1}/\textbf{28.2}/52.0/79.7/64.3/25.1   \\
		\whline
	\end{tabular}
	\label{Table-lb-loc}
\end{table*}

To intuitively understand the performance of crowd localization, Fig. \mbox{\ref{Fig-vis-loc}} demonstrates the visualization results of four methods on some typical samples. For the first sample, containing different-scale heads, Faster RCNN almost misses all small objects. The other three methods perform better for this scene than it. For large-scale objects (such as the second sample), Faster RCNN produces the perfect results of 100\% precision and 100\% recall. TinyFaces is the second, and the other two methods miss quite a few heads to different extents. In congested crowd scenes (e.g., Sample 3), VGG+GPR and RAZ\_Loc obtain good results though they produce some false positives in the background regions. Faster RCNN and TinyFaces work not well for this case: the former miss 95.4\% heads, and the latter yields many false positives. Besides, we also find TinyFaces produces more false positives than other methods.

In summary, there is no method to tackle the crowd localization problem well: 1) the traditional general object detection methods can not detect small-scale objects; 2) TinyFaces outputs quite a few false positives; 3) the existing regression-/classification methods can not handle large-range scale variations and mis-estimations on the background.

\subsection{Leaderboard}

Table \mbox{\ref{Table-lb-loc}} reports the results of four methods on \emph{the test set}. It lists the overall localization performance (F1-measure, Precision, and Recall) under $\sigma_l$ and counting performance (MAE, MSE, NAE), category-wise Recall on the different head scales (Box Level). Compared with the results of \emph{the validation set}, we find that the rankings are consistent. For the primary key (overall F1-measure), RAZ\_Loc attains the first place. From the category-wise results of Box Level, we find that all methods perform poorly for tiny heads (area range is $[1,10]$). The main reason is that the current feature extractor loses spatial information (usually, $8\!\times$ or $16\!\times$ downsampling are adopted). For extremely large-scale heads (area is more than $10^5$), detection-based methods is better than regression-/classification methods. The main reason is the latter's label is so small (VGG+GPR: $15\!\times\!15$, RAZ\_Loc: $3\!\times\!3$) that can not cover enough semantic head regions.

\section{Conclusion and Outlook}
\label{conclusion}

In this paper, a large-scale NWPU-Crowd counting dataset is constructed, which has the characteristics of high resolution, negative samples, and large appearance variation. At the same time, we develop an online benchmark website to fairly evaluate the performance of counting models. Based on the proposed dataset, we perform the fourteen typical algorithms and rank them from the perspective of the counting and localization performance, the density map quality, and the time complexity.

According to the quantitative and qualitative results, we find some interesting phenomena and some new problems that need to be addressed on the proposed dataset: 

\begin{enumerate}
	\item[1)] \textbf{How to improve the robustness of the models?} In the real world, the counters may encounter many unseen data, giving incorrect estimation for background regions. Thus, the performance on negative samples is vital in the counting, which represents the models' robustness.
	
	\item[2)] \textbf{How to remedy the performance impact between different scenes?} Due to the large appearance variations, the training with all data results in an obvious performance reduction compared with the individual training for each category. Hence, it is essential to prompt the counting model's capacity for appearance representations. 
	
	\item[3)] \textbf{How to reduce the estimation errors in the extremely congested crowd scenes?} Because of head occlusions, small objects, and lack of structured information, the existing models can not work well in the high-density regions. 
	
	\item[4)] \textbf{How to accurately locate the tiny-size and large-scale heads together?} The current detection-/regression-/classification-based methods can not handle the problem of large-range scale variation in real crowd scenes. Perhaps researchers need to design scale-aware models or hybrid methods to locate the head position accurately.
	
\end{enumerate}

In the future, we will continue to focus on handling the above issues and dedicate to improving the performance of crowd counting and localization in the real world.

{\small
\bibliographystyle{IEEEtran}
\bibliography{IEEEabrv,reference}
}

\end{document}